\documentclass[conference]{IEEEtran}
\IEEEoverridecommandlockouts
\usepackage{cite}
\usepackage{amsmath,amssymb,amsfonts}
\usepackage{algorithmic}
\usepackage{graphicx}
\usepackage{threeparttable}
\usepackage{textcomp}
\usepackage{svg}
\usepackage{xcolor}
\usepackage{tabularx}
\colorlet{shadecolor}{gray!40}
\usepackage{tcolorbox}  

\usepackage[T1]{fontenc}
\usepackage{graphicx}
\usepackage{booktabs}
\usepackage{multirow}
\usepackage{rotating}
\usepackage{amsmath}
\usepackage{float}

\usepackage[table]{xcolor}
\usepackage{threeparttable} 

\usepackage{hyperref}

\usepackage{bm}
\def\BibTeX{{\rm B\kern-.05em{\sc i\kern-.025em b}\kern-.08em
    T\kern-.1667em\lower.7ex\hbox{E}\kern-.125emX}}

\usepackage[table]{xcolor}
\usepackage{multirow}
\usepackage{xfp}
\usepackage{ifthen}

\definecolor{lightred}{RGB}{255,220,220}
\definecolor{mediumred}{RGB}{255,150,150}
\definecolor{darkred}{RGB}{255,100,100} 

\definecolor{lightblue}{RGB}{220,235,255}
\definecolor{mediumblue}{RGB}{150,200,255}
\definecolor{darkblue}{RGB}{100,150,255} 

\newcommand{\colorcell}[2]{%
  \edef\cellvalue{#1}%
  \edef\cellbase{#2}%
  \ifthenelse{\equal{\cellvalue}{-}}{%
    -%
  }{%
    \ifthenelse{\equal{\cellbase}{0}}{%
      \ifthenelse{\fpeval{\cellvalue > 0}=1}{\cellcolor{darkred}\cellvalue}{%
        \cellvalue%
      }%
    }{%
      \edef\cellratio{\fpeval{\cellvalue / \cellbase}}%
      \ifthenelse{\fpeval{\cellratio > 1.5}=1}{\cellcolor{darkred}\cellvalue}{%
        \ifthenelse{\fpeval{\cellratio > 1.2}=1}{\cellcolor{mediumred}\cellvalue}{%
          \ifthenelse{\fpeval{\cellratio > 1.0}=1}{\cellcolor{lightred}\cellvalue}{%
            \ifthenelse{\fpeval{\cellratio < 0.5}=1}{\cellcolor{darkblue}\cellvalue}{%
              \ifthenelse{\fpeval{\cellratio < 0.8}=1}{\cellcolor{mediumblue}\cellvalue}{%
                \ifthenelse{\fpeval{\cellratio < 1.0}=1}{\cellcolor{lightblue}\cellvalue}{%
                  \cellvalue%
                }%
              }%
            }%
          }%
        }%
      }%
    }%
  }%
}

\makeatletter
\newcommand{\linebreakand}{%
  \end{@IEEEauthorhalign}
  \hfill\mbox{}\par
  \mbox{}\hfill\begin{@IEEEauthorhalign}
}
\makeatother

\begin{document}


\title{ReviewGuard: Enhancing Deficient Peer Review Detection via LLM-Driven Data Augmentation}

\author{
\IEEEauthorblockN{Haoxuan Zhang}
\IEEEauthorblockA{University of North Texas\\
Denton, TX, USA\\
HaoxuanZhang@my.unt.edu}
\and

\IEEEauthorblockN{Ruochi Li}
\IEEEauthorblockA{North Carolina State University\\
Raleigh, NC, USA\\
rli14@ncsu.edu}
\and
\IEEEauthorblockN{Sarthak Shrestha}
\IEEEauthorblockA{University of North Texas\\
Denton, TX, USA\\
SarthakShrestha@my.unt.edu}
\linebreakand 
\IEEEauthorblockN{Shree Harshini Mamidala}
\IEEEauthorblockA{University of North Texas\\
Denton, TX, USA\\
ShreeHarshiniMamidala@my.unt.edu}
\and
\IEEEauthorblockN{Revanth Putta}
\IEEEauthorblockA{University of North Texas\\
Denton, TX, USA\\
RevanthPutta@my.unt.edu}
\and
\IEEEauthorblockN{Arka Krishan Aggarwal}
\IEEEauthorblockA{University of North Texas\\
Denton, TX, USA\\
ArkaKrishanAggarwal@my.unt.edu}

\linebreakand 
\IEEEauthorblockN{Ting Xiao\IEEEauthorrefmark{1}}
\IEEEauthorblockA{University of North Texas\\
Denton, TX, USA\\
Ting.Xiao@unt.edu}
\and
\IEEEauthorblockN{Junhua Ding}
\IEEEauthorblockA{University of North Texas\\
Denton, TX, USA\\
Junhua.Ding@unt.edu}
\and
\IEEEauthorblockN{Haihua Chen\IEEEauthorrefmark{1}}
\IEEEauthorblockA{University of North Texas\\
Denton, TX, USA\\
Haihua.Chen@unt.edu}
\thanks{\IEEEauthorrefmark{1}Corresponding authors}
}

\maketitle

\begin{abstract}
Peer review serves as the gatekeeper of science, yet the surge in submissions and widespread adoption of large language models (LLMs) in scholarly evaluation present unprecedented challenges. While recent work has focused on using LLMs to improve review efficiency, unchecked deficient reviews from both human experts and AI systems threaten to systematically undermine academic integrity. To address this issue, we introduce ReviewGuard, an automated system for detecting and categorizing deficient reviews through a four-stage LLM-driven framework: data collection from ICLR and NeurIPS on OpenReview, GPT-4.1 annotation with human validation, synthetic data augmentation yielding 6,634 papers with 24,657 real and 46,438 synthetic reviews, and fine-tuning of encoder-based models and open-source LLMs. Feature analysis reveals that deficient reviews exhibit lower rating scores, higher self-reported confidence, reduced structural complexity, and more negative sentiment than sufficient reviews. AI-generated text detection shows dramatic increases in AI-authored reviews since ChatGPT's emergence. Mixed training with synthetic and real data substantially improves detection performance—for example, Qwen 3-8B achieves recall of 0.6653 and F1 of 0.7073, up from 0.5499 and 0.5606 respectively. This study presents the first LLM-driven system for detecting deficient peer reviews, providing evidence to inform AI governance in peer review. Code, prompts, and data are available at \href{https://github.com/haoxuan-unt2024/ReviewGuard}{GitHub Repository}.
\end{abstract}


\section{Introduction}

Peer review serves as the cornerstone of scientific communication, ensuring the quality, validity, and integrity of research before publication~\cite{tennant2017multi,jin2024agentreview}. Through expert evaluation, it maintains stringent standards in methodology, precision, and scholarly contribution~\cite{yuan2022can}. However, the system faces significant pressures, with submission volumes growing exponentially, overwhelming review processes, and compromising quality at scale~\cite{wei2025ai}. Compounding factors include heavy reviewer workloads and compressed timelines that foster superficial evaluations, suboptimal reviewer-manuscript pairings, and reduced accountability due to anonymity~\cite{warne2016rewarding,willis2016peer,tomkins2017reviewer}.

Advancements in LLMs present viable solutions, as evidenced by studies showing that LLM-generated reviews often match human assessments, pinpoint methodological problems, evaluate contributions, and provide constructive feedback~\cite{yuan2022can,jin2024agentreview,zhang2025evolving}. Their capacity for rapid, consistent processing of extensive text supports scalable peer review enhancements~\cite{wei2025ai}. However, integrating LLMs introduces risks, such as prompt injection vulnerabilities that enable biased reviews, along with model-inherent biases, hallucinations, and generic outputs that undermine the depth and trustworthiness of the review~\cite{liang2024monitoring,wei2025ai,li2025unveiling}.

Existing research on LLMs in peer review predominantly emphasizes replicating or improving review practices, with limited focus on identifying and mitigating deficient reviews~\cite{li2025unveiling}. This oversight is concerning, as unchecked problematic behaviors from human or AI reviewers could exacerbate systemic flaws. Barriers to advancement include the lack of standardized datasets, metrics, and detection mechanisms, issues intensified by accelerating LLM adoption.

To address these challenges, we introduce ReviewGuard, an LLM-powered framework designed for the detection and analysis of deficient peer reviews. This methodology integrates peer review data acquisition from the OpenReview platform, focusing on top machine learning conferences like ICLR and NeurIPS, with a focus on manuscripts featuring conflicting assessments to produce an initial dataset of 24,657 authentic reviews from 6,634 papers. The annotation leverages GPT-4.1 for the preliminary labeling, validated by human experts, to categorize reviews as sufficient or deficient, including subtypes such as superficiality and lack of constructiveness, followed by an augmentation with 46,438 LLM-generated synthetic reviews to address class imbalances. Multi-dimensional feature analysis examines textual attributes, including structural metrics like complexity indices and qualitative elements such as sentiment profiles and AI content detection. Finally, detection models are developed via fine-tuning of encoders including BERT, SciBERT, and RoBERTa, as well as open LLMs like LLaMa 3.1-8B-Instruct and Qwen 3-8B, enabling automated identification of deficient review.

Our feature analysis reveals distinctive patterns that distinguish deficient reviews (DR) from sufficient reviews (SR).

\begin{itemize}
    \item \textbf{Structural Complexity}: SR demonstrates greater linguistic depth, with averages of 24.61 sentences and 425.47 lexical tokens, compared to 19.43 sentences and 318.44 tokens for DR, alongside higher complexity scores (e.g., Linsear Write readability of 14.02 versus 12.68).
    \item \textbf{Metadata Correlations}: SR is associated with elevated ratings (mean 5.37 versus 3.74 for DR; Spearman $\rho=0.256$, $p<0.001$) and slightly lower confidence (mean 3.68 versus 3.75; $\rho=-0.046$, $p<0.001$), reflecting balanced and reflective evaluations.
    \item \textbf{Sentiment Profiles}: SR is predominantly neutral ($\geq90\%$), while DR exhibit 4–5 times more negativity, indicating potential emotional bias ($\chi^2=397.7$ for ICLR and 29.1 for NeurIPS, both $p<0.001$).
    \item \textbf{AI-Generated Content Trends}: Detection via the Binoculars~\cite{hans2024spotting} framework reveals a post-ChatGPT surge in AI-written reviews, with over 100 AI-detected SR in ICLR 2025 and escalations to >40 in NeurIPS 2024, alongside rising DR detections, underscoring authenticity deficits. Notably, AI adoption is widespread across the spectrum of review quality, appearing in both sufficient and deficient reviews.

\end{itemize}

These patterns emphasize DR deficiencies in professionalism, depth, and genuineness, frequently linked to hasty or superficial judgments.

This work advances academic integrity through the following key contributions:

\begin{itemize}
    \item We propose \textbf{ReviewGuard}, a scalable, four-stage framework that integrates data curation, annotation, analytical scrutiny, and detection, meticulously tailored for machine learning conference contexts.
    \item Our analysis quantifies prevalent DR patterns, such as superficiality and lack of constructiveness, while highlighting the proliferation of AI-generated content, thereby informing policy frameworks for ethical LLMs integration in peer review.
    \item We improve data quality through LLM-driven data augmentation and demonstrate enhanced model efficacy, particularly in fine-grained multi-label detection of deficient subtypes.
\end{itemize}

ReviewGuard not only enables deficient review detection but also lays groundwork for robust, AI-enhanced peer review systems, preserving the reliability and fairness of scientific discourse.

\section{Related Work}

Peer review is fundamental to the reliability and integrity of academic publishing, serving as both a quality control mechanism and a platform for constructive academic exchange~\cite{lu2022research, august2024write}. Through critical evaluation of methodological soundness, originality, and clarity, reviewers help maintain disciplinary standards and prevent the dissemination of unreliable findings~\cite{kelly2014peer, glonti2019scoping}. Reviewers’ assessments not only determine whether research enters the scientific record, but also provide authors with expert feedback that strengthens their manuscripts prior to publication. The value of this process depends on the expertise, objectivity, and professionalism of reviewers, whose constructive and actionable feedback is essential for improving both the quality of individual manuscripts and the overall standard of scientific literature.

Despite its essential role, the peer review process continues to encounter challenges, including reviewer bias, inconsistency, limited coverage of important evaluation criteria, and an increasing burden on the research community \cite{langford2015arbitrariness}. Manual reviews are often time-consuming and costly, and numerous studies have highlighted common issues such as superficial assessments, insufficient justification, and unconstructive or careless feedback \cite{tomkins2017reviewer}. As the volume of submissions grows exponentially, these limitations have become more pronounced, raising pressing concerns about the overall quality and fairness of scientific evaluation. Recent advances in LLMs have created new opportunities to address some of these challenges. Some research has explored automated multi-agent systems that leverage LLMs to simulate and streamline the peer review process, demonstrating promising potential for improving scalability and efficiency \cite{jin2024agentreview, baek2024researchagent, ghafarollahi2025sciagents}. Meanwhile, hierarchical and structured frameworks have been proposed to address the limitations of superficial or unsubstantiated feedback by decomposing the review process into fine-grained, aspect-specific tasks and modeling the logical relationships between papers and review comments \cite{chang2025treereview, yu2024seagraph, yu2024automated}. Other studies have investigated augmentation strategies such as retrieval-augmented generation (RAG) and fine-tuning to further enhance review quality. RAG integrates targeted literature retrieval into the generation process to ground assessments in verifiable evidence \cite{liu2025improving}, while fine-tuning adapts LLMs to the stylistic and evaluative norms of peer review, improving factual accuracy and alignment with domain-specific criteria \cite{ yuan2022can}.

However, neither human reviewers nor LLM-based systems are immune to deficient reviewing. For human reviewers, such behavior often stems from constraints such as limited time, high workload, and gaps in domain expertise, which can lead to superficial assessments or insufficiently justified recommendations \cite{benos2007ups}. LLM-generated reviews may suffer from deep-level understanding, rely on repetitive template-like language, and rarely pose clarifying questions\cite{yuan2022can}. Given the increasing volume of submissions and the growing use of LLM-assisted review, addressing deficient review has become an urgent priority to safeguarding the integrity of scientific publication. 

\section{Methodology}

\begin{figure*}[htbp]
\centering
\includegraphics[width=\linewidth]{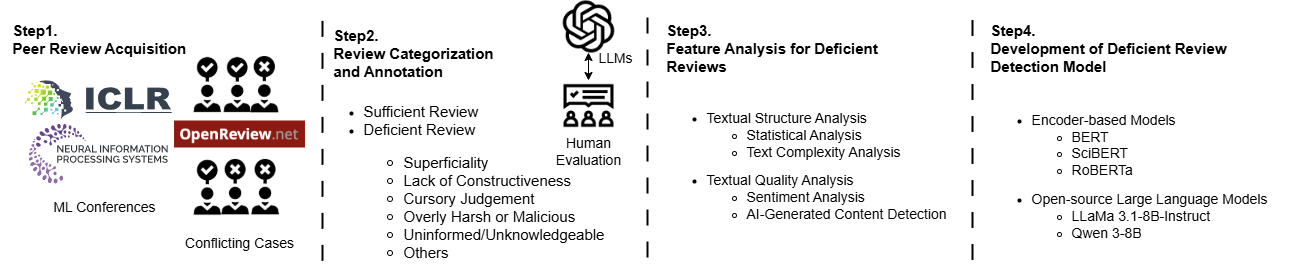}
\caption{The ReviewGuard pipeline for automated deficient review detection. The framework integrates four sequential stages: (1) targeted data acquisition from OpenReview focusing on conflicting reviews, (2) LLM-driven annotation with human validation and synthetic data augmentation to address class imbalance, (3) multi-dimensional feature analysis examining textual structure and quality, and (4) fine-tuning of encoder-based and open-source LLM detection models.}
\label{framework}
\end{figure*}

The ReviewGuard pipeline (as shown in Figure~\ref{framework}) employs a systematic four-stage approach for detecting deficient peer reviews in machine learning conferences. It begins with targeted acquisition of review texts from OpenReview for ICLR and NeurIPS, prioritizing submissions with conflicting assessments to capture edge cases. The second stage involves GPT-4.1–assisted, human-validated annotation, labeling reviews as sufficient or deficient and assigning fine-grained deficient codes (e.g., superficiality, lack of constructiveness, cursory judgment, excessive harshness, malice, or uninformed commentary). This is augmented by synthetic review generation using LLMs to simulate diverse sufficient and deficient archetypes, enriching the dataset. The third stage performs feature analysis on both real and synthetic deficient reviews, examining textual structure (statistical and complexity metrics) and quality (sentiment and AI-generated content detection). Finally, detection models are developed through domain-adaptive fine-tuning of encoder-based models (BERT, SciBERT, RoBERTa) and low-resource fine-tuning of open-source LLMs (LLaMa 3.1-8B-Instruct, Qwen 3-8B), enabling automated identification of deficient content.

\subsection{Peer Review Acquisition}

OpenReview proactively publishes manuscripts and peer reviews, enhancing transparency in scientific communication~\cite{sun2025openreview}. ICLR and NeurIPS review data are particularly valuable because these premier machine learning conferences directly impact academic integrity, and both publish comprehensive review artifacts for accepted and rejected submissions, ensuring data representativeness and mitigating selection bias~\cite{wang2024effective,wang2024content}. OpenReview's API enables systematic, large-scale acquisition supporting reproducibility and exhaustive coverage.

Deficient reviews constitute a minor proportion of the total corpus. To maximize detection within a feasible annotation budget, we focus on manuscripts with conflicting peer review outcomes. Conflicting reviews potentially indicate deficient reviews, as substantial divergence may reflect superficial engagement, insufficient domain expertise, or inconsistent evaluation criteria. Extreme outlier opinions may signal inadequate engagement rather than legitimate scholarly disagreement, warranting scrutiny for deficiencies in comprehension, substantiation, or constructive feedback.

We operationalize ``conflicting reviews'' as manuscripts where one reviewer's assessment diverges substantially from the consensus. For each manuscript with $n$ reviews, we compute a consensus score by averaging all ratings after excluding the maximum and minimum values. The absolute differences between the extreme scores and consensus, denoted $\mathit{diff}_{\text{high}}$ and $\mathit{diff}_{\text{low}}$, quantify the extent of divergence. This selection criterion enriches the dataset with reviews more likely to exhibit deficiency patterns while maintaining representation of both sufficient and deficient reviews.

\subsection{Review Categorization and Annotation}
\label{s3.2}

\subsubsection{Definitions}

Deficient peer review is not accidental. It grows out of systemic and personal pressures that now shape academic publishing \cite{tennant2017multi}. First, reviewers often face heavy workloads and tight deadlines, leading to short and shallow evaluations. Because reviewing is voluntary and rarely rewarded in formal evaluations, good intentions can give way to minimal effort \cite{warne2016rewarding,willis2016peer}. Second, the surge in submissions and the imperfect matching of papers with reviewers mean that some manuscripts are judged by people who lack the required expertise \cite{yuan2022can}. Third, closed and anonymous review systems offer little accountability, lowering the cost of harsh or unsupported criticism and even allowing reviewers to harm competing work \cite{tomkins2017reviewer}. Finally, most journals and conferences do not provide formal training, so new reviewers may rely on vague, template-like comments, mistaking length for depth~\cite{august2024write}. Together, these conditions make deficient reviews much more likely.

Review responsibility can be conceptualized along three interdependent dimensions \cite{yuan2022can,jin2024agentreview}: \emph{commitment} (the extent of substantive engagement with the manuscript), \emph{constructive intention} (a fair-minded desire to improve the work) and \emph{domain knowledgeability} (sufficient expertise to make an informed judgment). A report that satisfies all three pillars yields what we term a \textit{sufficient review (SR)}; editors and authors can rely on such feedback to reach defensible editorial decisions. A breakdown in any single dimension produces a \textit{deficient review (DR)}: the report ceases to perform its gate-keeping or mentoring functions and may distort acceptance outcomes or discourage author participation. The first task in categorization, therefore, is to determine whether the text demonstrates substantive engagement, benign intent, and adequate expertise. When all criteria are met the review is coded as sufficient; failure on one or more dimensions triggers a deficient label that is then sub-typed according to its dominant failure mode.

Deficient reviews exhibit recurring patterns that can be categorized into six types based on empirical evidence \cite{yuan2022can,jin2024agentreview,sizo2025defining}. These include \emph{superficial} comments, which are shallow or vague; those \emph{lacking constructiveness}, which list faults without offering actionable remedies; and \emph{cursory judgments}, which reveal minimal effort and often misquote or ignore key sections. Some reviews are \emph{overly harsh or malicious}, dismissing the paper without balanced reasoning, while others are simply \emph{uninformed}, containing factual errors that indicate limited expertise. Finally, the \emph{others} category captures additional deviations, such as prestige, nationality, or gender bias, that manifest deficiency not covered by the preceding types. Table~\ref{tab:definitions} formalizes the distinction between sufficient and deficient reviews and summarizes these six subtypes of deficient review.

\begin{table}[htbp]
\centering
\caption{Responsibility categories and their operational definitions.}
\label{tab:definitions}
\begin{tabular}{p{0.25\linewidth}p{0.65\linewidth}}
\toprule
\textbf{Category} & \textbf{Definition}\\ \midrule
Sufficient review (SR) & Meets all three pillars—commitment, constructive intention, and domain knowledgeability—thus providing substantive, fair, and actionable feedback.\\
Deficient review (DR) & Fails at least one pillar above; the prevailing failure mode is further coded into one of the sub-types listed below.\\
\midrule
\multicolumn{2}{l}{\emph{Deficient review: sub-types}}\\
\midrule
Superficiality & Comments are shallow, vague, or perfunctory, showing minimal engagement with the manuscript.\\
Lack of constructiveness & Feedback notes problems but offers no actionable guidance for improvement.\\
Cursory judgment & Review appears hastily written, misquotes content, or otherwise signals the paper was not thoroughly read.\\
Overly harsh / malicious & Tone is unnecessarily negative or dismissive without balanced reasoning.\\
Uninformed / unknowledgeable & Contains factual errors or requests reflecting insufficient expertise in the paper’s area.\\
Others & Any additional deficient (prestige, nationality, or gender bias) not captured above.\\
\bottomrule
\end{tabular}
\end{table}

\subsubsection{Annotation Protocol}

Large-scale data annotation is inherently time-consuming and labor-intensive. However, rapid advancements in LLMs have placed them as effective annotators, widely adopted in various disciplines \cite{tan2024large}. Empirical evidence indicates that LLM-generated annotations often achieve or surpass the quality of human annotations, with inter-LLMs consistency exceeding that observed among human annotators~\cite{he2024annollm}. However, LLMs are not infallible oracles; their outputs require further human validation to ensure reliability. To address this, we integrate LLM-driven large-scale annotation with human verification on a smaller scale \cite{wang2024human}. Specifically, we employ the GPT-4.1 model \footnote{GPT-4.1: \url{https://openai.com/index/gpt-4-1/}.} for annotation. The input comprises the title, abstract, and full text and metadata of all associated reviews. The model is tasked with determining whether each review qualifies as sufficient; If considered deficient, it categorizes the review according to the six types outlined in Table~\ref{tab:definitions} and provides a rationale for classification. 


To rigorously assess the reliability of LLM-based annotation for subsequent analysis, we conduct a two-stage validation protocol. After initial labeling by GPT-4.1, a sample of reviews is independently annotated by expert human raters. We then compute Cohen's kappa \cite{carletta1996assessing} and Fleiss's kappa \cite{falotico2015fleiss} to quantify inter-rater agreement on both responsibility labels and subtypes.

\subsubsection{Synthetic Review Augmentation}

The review data collected from the OpenReview platform exhibits significant data imbalance and low-resource characteristics. Utilizing LLMs for synthetic data generation has emerged as an effective approach for improving dataset quality and enhancing model performance \cite{feng2021survey,yu2023large}. This technology has been widely applied across multiple domains within natural language processing, including classification tasks, summarization, question-answering, and translation, demonstrating promising results~\cite{feng2021survey,benayas2024enhancing}. Therefore, we adopt synthetic data augmentation techniques to address the data scarcity problem.

Specifically, we design specialized prompts targeting different types of deficient review behaviors and generate synthetic review data through LLMs based on paper information. The prompts are based on Table~\ref{tab:definitions}. For each paper identified as a conflicting case, we generate six distinct types of deficient review samples and one sufficient review to ensure data diversity and completeness.

\subsection{Feature Analysis for Deficient Reviews}

We conduct an analysis on the real review dataset annotated by GPT-4.1 to reveal key differences between sufficient and deficient reviews. This analysis covers textual structure and textual quality aspects. Textual structure analysis includes descriptive statistics and complexity evaluation. Descriptive statistics explore how review type (sufficient vs. deficient) affects outcomes like review scores and self-reported reviewer confidence, demonstrating whether deficient reviews harm overall quality and quantifying any biases in the evaluation process. Textual complexity analysis highlights linguistic differences at the word and sentence levels, showing how review type influences structural sophistication and readability. Textual quality analysis focuses on sentiment analysis and AI-generated content detection, examining the emotional tone and authenticity of the reviews. Recently, the rise of LLMs has led to AI-generated reviews—a hard-to-detect form of academic misconduct that erodes the integrity and respect of scholarly peer review. 

\subsubsection{Textual Structure Analysis}

We employ Spearman's rank correlation coefficient—a widely used statistical method in quantitative research—to validate the relationships between review responsibility and both review scores and confidence scores. For textual complexity analysis, we utilize Textstat\footnote{Textstat: \url{https://github.com/textstat/textstat}.} to compute complexity metrics at the word, syllable, and sentence levels. This tool is selected for its validated algorithms, which provide objective, reproducible measures of linguistic intricacy, thereby substantiating differences in cognitive effort between sufficient and deficient reviews through quantifiable proxies of textual sophistication.

\subsubsection{Textual Quality Analysis}

We implement sentiment analysis using the twitter-roberta-base-sentiment-latest model \cite{camacho2022tweetnlp} to classify the emotional tone of the review texts, categorizing them into positive, negative, or neutral sentiments to reveal potential biases or inconsistencies in sufficient versus deficient reviews. For AI-generated content detection, we implement the state-of-the-art Binoculars \cite{hans2024spotting} framework, which leverages the Falcon-7B and Falcon-7B-Instruct models to achieve high accuracy in zero-shot detection without requiring pre-training. This method is justified by its demonstrated superiority in recent benchmarks for distinguishing human-authored from AI-generated text, relying on cross-model perplexity comparisons that exploit inherent distributional differences, thus providing a reliable safeguard against undetected academic impropriety in review datasets.

\subsection{Development of Deficient Review Detection Model}

To develop a deployable system for the identification and classification of deficient peer reviews, we examine the viability of fine-tuning low-resource, open-source language models. Although closed-source commercial models exhibit robust zero-shot capabilities, they present significant limitations, including elevated inference costs, constrained reproducibility, opacity in training data that may introduce undisclosed biases, and restrictive licensing that hampers academic dissemination and examination. We thus fine-tune a curated selection of publicly accessible models on our dataset. The selected models are enumerated below:

\begin{itemize}
\item \textbf{BERT-base-uncased} \cite{devlin2019bert}: Serves as a foundational model for sequence modeling, offering a standardized reference for evaluating enhancements in natural language processing tasks.
\item \textbf{SciBERT-scivocab-uncased} \cite{beltagy2019scibert}: Pre-trained on a corpus of 1.14 million scientific abstracts, it provides domain-adapted embeddings and vocabulary optimized for technical discourse in machine learning and related fields.
\item \textbf{RoBERTa-base} \cite{liu2019roberta}: An advanced refinement of BERT, incorporating extended training durations, dynamic masking, and increased batch sizes to achieve enhanced robustness and generalization across diverse textual datasets.
\item \textbf{LLaMa 3.1-8B-Instruct} \cite{dubey2024llama}: An 8-billion-parameter model fine-tuned for instruction adherence, striking an optimal balance between generative efficacy and computational parsimony.
\item \textbf{Qwen 3-8B} \cite{yang2025qwen3}: A similarly scaled 8-billion-parameter model, notable for its multilingual pre-training and proficient handling of instructional prompts.
\end{itemize}

\section{Experiments and Results}

\subsection{Experimental Setup}

To promote reproducibility, we detail the specific experimental environment and parameter configurations below. For BERT, SciBERT and RoBERTa, the maximum token limit was set to 512, with 10 training epochs and the incorporation of an early stopping mechanism. For open-source LLMs, we employed the LoRA fine-tuning technique \cite{hulora} with the following parameters: $r=8$, $\alpha=32$, and LoRA dropout$=0.05$. Additionally, BF16 precision and Flash Attention 2 were utilized, with a maximum sequence length of 6144, 1 training epoch, and a learning rate of $1 \times 10^{-4}$. The closed-source language model versions used was gpt-4.1-2025-04-14, configured with a temperature of 0.85 and top\_p$=1.0$.

\subsection{Dataset Preparation}
\subsubsection{Initial Data Curation}
\begin{figure}[htbp]
\centering
\includegraphics[width=\linewidth]{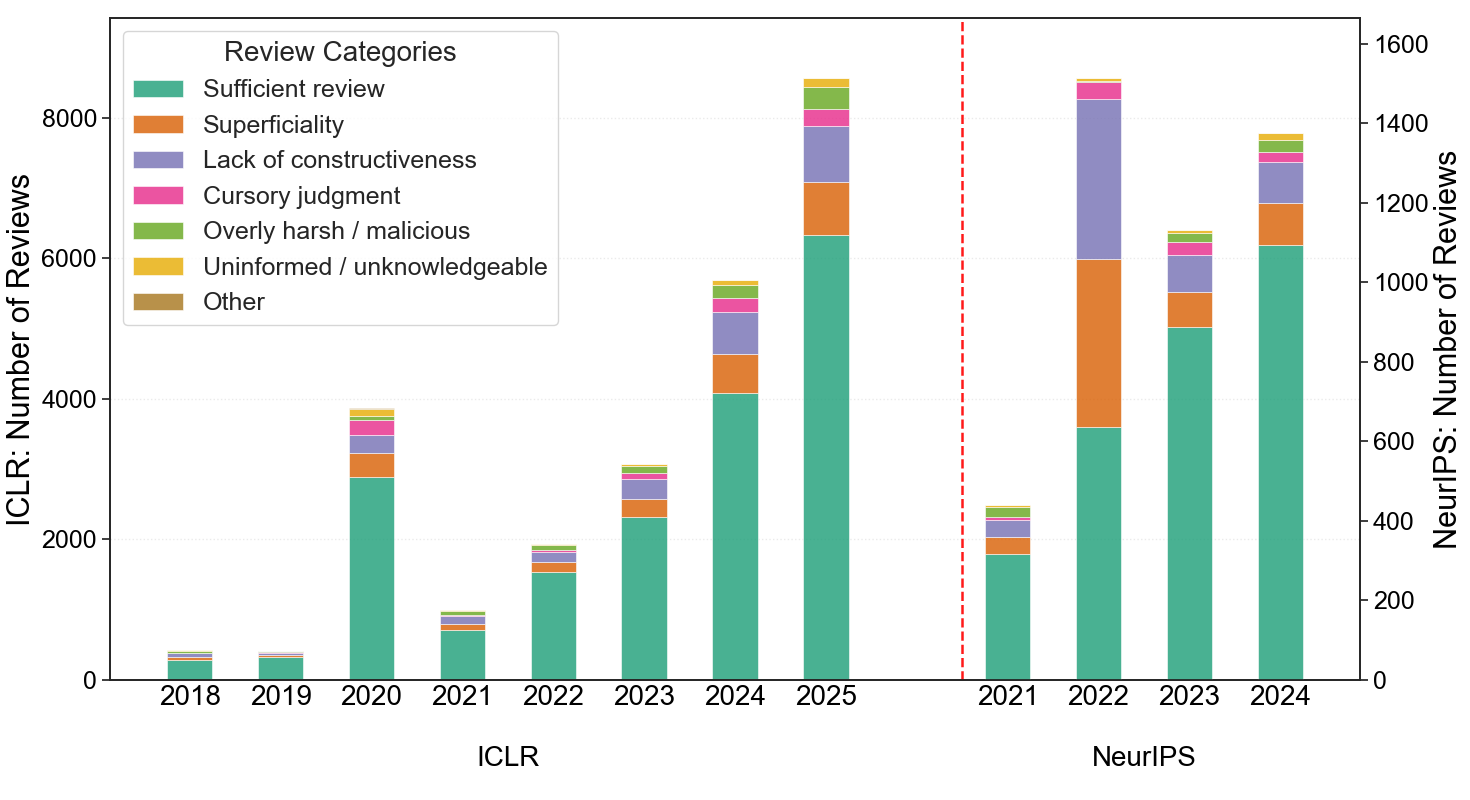}
\caption{Distribution of review categories in ICLR and NeurIPS conferences.}
\label{review-dis}
\end{figure}

We utilized the OpenReview API to collect titles, abstracts, and corresponding review information for papers from ICLR 2018--2025 and NeurIPS 2021--2024, with a data cutoff date of June 2025. To maximize the identification of potential deficient reviews, we applied the controversial paper identification formula outlined in the Section~\ref{s3.2}, thereby detecting 14.56\% (6,634) of papers exhibiting a score divergence of at least 3, which yielded a total of 24,657 review reports. Of these, 86.89\% (21,425) were labeled as sufficient reviews, while 13. 11\% (3,232) were classified as deficient (examples of these reviews are provided in Table~\ref{tab:academic-reviews}). Evidently, the data set suffers from severe imbalance, characterized by an under-representation of deficient reviews. Furthermore, we conducted a distribution analysis of the review categories across the two conferences, as depicted in Figure~\ref{review-dis}. The findings reveal that \textit{lack of constructiveness} and \textit{superficiality} are the predominant types of deficient reviews, followed by \textit{overly harsh or malicious} comments and \textit{cursory judgments}, with \textit{uninformed} reviews and the \textit{others} category comprising the smallest shares. This pattern indicates that the deficient reviews at ICLR and NeurIPS primarily fail to provide detailed or actionable feedback, thereby introducing the risks of emotionally charged or hasty evaluations in the review process.

\begin{table*}[htbp]
\centering
\caption{Examples of Deficient Reviews and Associated Categories from OpenReview}
\label{tab:academic-reviews}
\begin{tabular}{p{0.9cm}p{12cm}p{3.7cm}}
\toprule
\textbf{Review Link} & \textbf{Review Content (Excerpt)} & \textbf{Category (Annotated by GPT-4.1)} \\
\midrule
\href{https://openreview.net/forum?id=z2VBHpRT14&noteId=vxtHpMxjZ7}{Link} & Given my limited expertise in the field of space observations, I find myself ill-equipped to provide a comprehensive evaluation of this paper. The specific challenges and nuances within this domain are not within my area of specialization. Therefore, I recommend that you consult with a diverse group of reviewers who possess a deeper understanding of space-related research to ensure a thorough and informed assessment of the paper's content and significance. & Cursory judgment \newline Uninformed / unknowledgeable \\
\hline
\href{https://openreview.net/forum?id=zFWKKYz2yn&noteId=I1qFXXjYvy}{Link} & The paper is well presented, by providing substantial background information and experiments. … Honestly, my familiarity with current RNN research, including rule extraction strategies, is limited. & Lack of constructiveness \newline Overly harsh / malicious \\
\hline
\href{https://openreview.net/forum?id=x5txICnnjC&noteId=XVDPQVLycx}{Link} & The main drawback is the multitude of assumptions made in the article, leading to a highly robust conclusion. … And the conclusion itself lacks significance. What matters is, assuming the article's conclusion is correct, why we would choose this specific geometry, what advantages it offers, and how we arrive at such a choice. & Lack of constructiveness \newline Overly harsh / malicious \newline Uninformed / unknowledgeable \\
\hline
\href{https://openreview.net/forum?id=yJdj2QQCUB&noteId=7kd1xPqDvZ}{Link} & … Could you please verify whether the owner of the Google Drive link is the author? (if you want to check the name that I found, then I will leave the name in the comment) … & Others \\
\hline
\href{https://openreview.net/forum?id=B9qg3wo75g&noteId=SD4DIWoBW1}{Link} & … If the authors really want to make some contributions to image generation, I suggest they spend more energy understanding the structure of image data, instead of focusing only on plugging in a fractional BM to replace BM. … & Superficiality \newline Lack of constructiveness \newline Cursory judgment \newline Overly harsh / malicious \\
\bottomrule
\end{tabular}
\end{table*}

\subsubsection{Data Augmentation}
To mitigate the severe imbalance in the dataset, we applied synthetic data augmentation by generating six distinct deficient reviews and one sufficient review for each paper, yielding a total of 46,438 augmented reports. Detailed statistics for the dataset are provided in Table~\ref{tab:dataset-statistics}.

\begin{table}[htbp]
\centering
\caption{Dataset Statistics in ICLR and NeurIPS Conferences}
\label{tab:dataset-statistics}
\resizebox{\linewidth}{!}{
\begin{tabular}{p{1.1cm}p{0.9cm}p{1.7cm}p{0.7cm}p{1cm}}
\toprule
\textbf{Venue} & \textbf{\#Papers} & \textbf{Conflicting} & \textbf{Real} & \textbf{Synthetic} \\
\midrule
ICLR'18 & 922 & 112 (12.1\%) & 338 & 784 \\
ICLR'19 & 1,419 & 118 (8.3\%) & 362 & 826 \\
ICLR'20 & 2,213 & 1,088 (49.2\%) & 3,272 & 7,616 \\
ICLR'21 & 2,594 & 213 (8.2\%) & 826 & 1,491 \\
ICLR'22 & 2,617 & 447 (17.1\%) & 1,697 & 3,129 \\
ICLR'23 & 3,793 & 690 (18.2\%) & 2,612 & 4,830 \\
ICLR'24 & 7,262 & 1,219 (16.8\%) & 4,720 & 8,533 \\
ICLR'25 & 11,520 & 1,759 (15.3\%) & 7,207 & 12,313 \\
\midrule
NeurIPS'21 & 2,768 & 96 (3.5\%) & 363 & 672 \\
NeurIPS'22 & 2,824 & 295 (10.4\%) & 1,061 & 2,065 \\
NeurIPS'23 & 3,395 & 289 (8.5\%) & 985 & 2,023 \\
NeurIPS'24 & 4,236 & 308 (7.3\%) & 1,214 & 2,156 \\
\midrule
\textbf{Total}  & 45,563 & 6,634 (14.6\%) & 24,657 & 46,438 \\
\bottomrule
\end{tabular}
}
\vspace{0.2cm} 
\footnotesize

\textit{Note:} Conflicting cases are defined as those where $\mathit{diff_{high}} \lor \mathit{diff_{low}} \geq 3$. ``\#Papers'' (column 2) indicates the number of submissions with sufficient reviews to compute consensus (due to many authors choosing to withdraw, these figures do not represent the actual submission numbers). ``Conflicting'' (column 3) provides the count and percentage of cases exceeding the threshold. ``Real'' (column 4) and ``Synthetic'' (column 5) show the counts of real and synthetic reviews, respectively.
\end{table}

\subsubsection{Human Evaluation}

To assess GPT-4.1's annotation accuracy, we conducted human evaluation with two evaluators (one undergraduate and one graduate student, both with computer science and data science expertise). Evaluators received systematic training on task objectives, evaluation criteria, and procedures, then independently assessed all samples without communication to ensure objectivity.

In the first round, we randomly selected 100 reviews for manual assessment, achieving Cohen's kappa of 0.4899 and Fleiss' kappa of 0.4740. Analysis of inconsistent cases revealed three primary disagreement sources: (1) boundaries between ``superficial'' and ``cursory judgment'' categories, (2) thresholds distinguishing ``overly harsh'' from constructively critical feedback, and (3) interpretation of domain-specific terminology for ``uninformed'' reviews. We refined the annotation guidelines and GPT-4.1 prompt by adding concrete examples, establishing clearer decision rules, and developing a consensus framework for borderline cases.

In the second round, evaluators independently assessed another 100 reviews using updated criteria, achieving Cohen's kappa of 0.8774 and Fleiss' kappa of 0.8793. These results confirm substantial improvement in inter-rater agreement and validate GPT-4.1's annotation quality, establishing the labeled data as a reliable foundation for subsequent analyses.

\subsection{Empirical Results from Feature Analysis of Deficient Reviews}
\subsubsection{Textual Structure Analysis}
To investigate the differences in textual structure between sufficient and deficient reviews, we computed Spearman correlation coefficients between review types and review metadata (rating scores and self-reported confidence) as well as text complexity metrics, as presented in Table~\ref{tab-corr}.

\begin{table}[htbp]
\centering
\caption{Correlation and Mean Statistics for Reviews.}
\label{tab-corr}
\begin{tabular}{lcc}
\toprule
 \textbf{Metric} & \textbf{Mean (SR vs. DR)} & \textbf{Correlation} \\ 
\midrule
rating score    & 5.37 (3.74)  & 0.256 *** \\
confidence score & 3.68 (3.75)  & -0.046 *** \\
linsear write formula &14.02 (12.68) &0.112 *** \\
sentence count & 24.61 (19.43)   & 0.151 *** \\
lexicon count  & 425.47 (318.44)   & 0.181 *** \\
syllable count  &753.42 (558.63)    & 0.188 *** \\
difficult words   &106.44 (79.54)  & 0.196 *** \\
monosyllable count &237.71 (180.01)  & 0.168 *** \\
\bottomrule
\end{tabular}
\vspace{0.2cm}
\footnotesize

\textit{Note:} Correlations shown are between the binary SR indicator (1=Sufficient Review, 0=Deficient Review) and each metric, with significance levels (*** indicates $p < 0.001$). The ``Mean'' column shows average values for SR and DR respectively. Higher positive correlations indicate metrics that are elevated in sufficient reviews compared to deficient reviews.
\end{table}

Table~\ref{tab-corr} highlights two metadata signals that most clearly separate sufficient reviews from deficient reviews.

\begin{itemize}
  \item \textbf{Rating--Review Type Link}. A moderate positive correlation between being a sufficient review and the overall rating (\(\rho = 0.256\), \(p < 0.001\)) indicates that sufficient reviews tend to receive or assign higher scores on average; the mean rating increases from 3.74 for deficient (DR) to 5.37 for sufficient (SR).
  \item \textbf{Confidence--Review Type Trade-off}. The SR indicator shows a weak but significant negative correlation with self-reported confidence (\(\rho = -0.046\), \(p < 0.001\)); the means differ slightly (3.75 for DR vs. 3.68 for SR), suggesting sufficient reviews exhibit marginally lower expressed certainty. 
\end{itemize}

Taken together, higher ratings coupled with slightly lower confidence characterize sufficient reviews as constructive, merit-recognizing, and self-reflective, in contrast to the harsher, more self-assured tone often observed in deficient reviews.

In terms of textual complexity, Table~\ref{tab-corr} shows that SR consistently exhibit greater linguistic sophistication than DR. Key observations are summarized below.

\begin{itemize}
\item \textbf{Global complexity}. SR is positively associated with several reported complexity metrics: Linsear-Write readability ($\rho = 0.112$), sentence count ($\rho = 0.151$), lexicon count ($\rho = 0.181$), syllable count ($\rho = 0.188$), difficult words count ($\rho = 0.196$), and monosyllable count ($\rho = 0.168$). Mean values reinforce this trend; for example, SR contains an average of 24.61 sentences and 425.47 lexical tokens, compared to 19.43 sentences and 318.44 tokens for DR.
\item \textbf{Cognitive demand}. The mean Linsear-Write score for SR is 14.02, corresponding to university-level readability, while the DR score is 12.68. This higher score suggests that SR employs discipline-specific terminology and more complex syntax, tailored to an expert audience.
\item \textbf{Elaboration versus brevity}. Elevated lexicon, syllable, and difficult words counts indicate that SR provides extensive, nuanced commentary, whereas the lower counts for DR suggest terse, potentially cursory feedback.
\end{itemize}
Collectively, these findings demonstrate that SR are marked by lexical richness, syntactic maturity, and detailed exposition, whereas DR tend to be shorter and linguistically simpler, thereby diminishing their corrective value in the scholarly evaluation process.

\subsubsection{Textual Quality Analysis}
\begin{figure}[htbp]
\centering
\includegraphics[width=\linewidth]{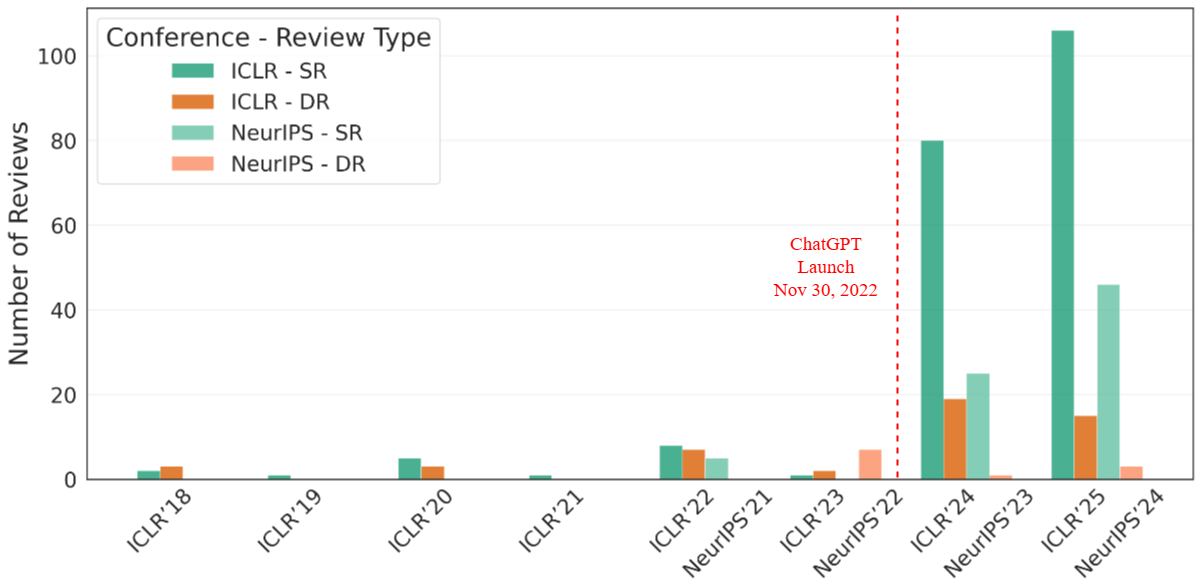}
\caption{Temporal analysis of AI-generated review detection using Binoculars framework in ICLR and NeurIPS conferences.}
\label{review-ai}
\end{figure}

We aim to evaluate differences in review quality between DR and SR by applying sentiment analysis and AI-generated content detection. We used the twitter-roberta-base-sentiment-latest model \cite{camacho2022tweetnlp} to identify positive, neutral, and negative sentiments within review texts. Furthermore, we applied $\chi^{2}$ tests to compare class frequencies, with results presented in Table~\ref{tab:sentiment_distribution}. We used the Binocular \cite{hans2024spotting} framework to detect AI-generated content in review texts, as shown in Figure~\ref{review-ai}.

\begin{table}[htbp]
\centering
\caption{Sentiment distribution (in percentages) of ICLR and NeurIPS reviews.}
\label{tab:sentiment_distribution}
\begin{tabular}{lccccccc}
\toprule
\textbf{Conference} & \multicolumn{3}{c}{\textbf{SR}} & \multicolumn{3}{c}{\textbf{DR}} \\
& \textbf{NEG} & \textbf{NEU} & \textbf{POS} & \textbf{NEG} & \textbf{NEU} & \textbf{POS} \\
\midrule
ICLR    & 0.49 & 90.21 & 9.30 & 4.60 & 88.37 & 7.03 \\
NeurIPS & 0.48 & 92.41 & 7.12 & 2.48 & 88.65 & 8.88 \\
\bottomrule
\end{tabular}
\vspace{0.2cm}
\footnotesize

\textit{Note:}
ICLR: $\chi^{2}$=397.7***, p < 0.001;  
NeurIPS: $\chi^{2}$=29.1***, p < 0.001.  
\end{table}

Based on the evidence from Table~\ref{tab:sentiment_distribution} and  Figure~\ref{review-ai}, we find several key observations:

\begin{itemize}
\item \textbf{DR and SR exhibit significantly different emotional patterns:} $\chi^{2}$ tests reveal substantial differences in sentiment between DR and SR for both ICLR ($\chi^{2}$ = 397.7***) and NeurIPS ($\chi^{2}$ = 29.1***). SR is overwhelmingly neutral in tone ($\geq$90\%), with extremely low negativity ($\leq$0.5\%). In contrast, DR contains less neutrality and 4--5 times higher negativity, indicating a less professional and more emotionally charged style.

\item \textbf{AI-generated reviews show clear temporal correlation with ChatGPT availability:} Figure~\ref{review-ai} demonstrates a pronounced increase in AI-detected reviews following the public launch of ChatGPT. For subsequent ICLR and NeurIPS cycles: ICLR 2024: SR $\approx$ 80, DR $\approx$ 20; ICLR 2025: SR $>$ 100, DR slightly declined. NeurIPS 2023: SR $>$ 20, DR single digits; NeurIPS 2024: SR $>$ 40, DR mildly increased. This temporal shift directly correlates with the widespread adoption of generative tools.

\item \textbf{AI adoption transcends reviewer quality boundaries:} AI-assisted review writing has become a widespread phenomenon, irrespective of reviewer responsibility classification. The higher absolute count of AI-generated reviews among SR is primarily due to SR forming the overwhelming majority of all reviews, rather than a disproportionate tendency among sufficient reviewers to use AI tools.
\end{itemize}

\begin{table*}[!ht]
\centering
\caption{Model Performance Comparison: Macro Average Metrics for Binary-SR/DR and Multi-DR Classification Tasks}
\label{tab:model_performance}
\small
\begin{tabular}{p{1.4cm} >{\centering\arraybackslash}p{1.2cm} >{\centering\arraybackslash}p{1.2cm} >{\centering\arraybackslash}p{1.2cm} >{\centering\arraybackslash}p{1.2cm} >{\centering\arraybackslash}p{1.5cm} >{\centering\arraybackslash}p{1.2cm} >{\centering\arraybackslash}p{1.2cm} >{\centering\arraybackslash}p{1.2cm} >{\centering\arraybackslash}p{1.5cm}}
\toprule
\multirow{2}{*}{\textbf{Model}} & \multirow{2}{*}{\textbf{Metric}} & \multicolumn{4}{c}{\textbf{Binary-SR/DR}} & \multicolumn{4}{c}{\textbf{Multi-DR}} \\
\cmidrule(lr){3-6} \cmidrule(lr){7-10}
 &  & \textbf{R, R} & \textbf{R, S} & \textbf{R+S, R} & \textbf{R+S, R+S} & \textbf{R, R} & \textbf{R, S} & \textbf{R+S, R} & \textbf{R+S, R+S} \\
\midrule
\multirow{3}{*}{\parbox{2cm}{BERT}}  
    & Precision & 0.4357 & 0.6211 & \colorcell{0.7485}{0.4357} & \colorcell{0.9495}{0.4357} & 0 & 0.0360 & \colorcell{0.5735}{0} & \colorcell{0.9596}{0} \\
    & Recall & 0.5000 & 0.7407 & \colorcell{0.6028}{0.5000} & \colorcell{0.9600}{0.5000} & 0 & 0.1052 & \colorcell{0.1347}{0} & \colorcell{0.8313}{0} \\
    & F1 score & 0.4656 & 0.5236 & \colorcell{0.6327}{0.4656} & \colorcell{0.9538}{0.4656} & 0 & 0.0536 & \colorcell{0.2132}{0} & \colorcell{0.8894}{0} \\
\midrule
\multirow{3}{*}{\parbox{2cm}{SciBERT}}  
    & Precision & 0.7785 & \textbf{0.6304} & \colorcell{0.7649}{0.7785} & \textbf{\colorcell{0.9520}{0.7785}} & 0.3047 & 0.0230 & \textbf{\colorcell{0.7809}{0.3047}} & \colorcell{0.9680}{0.3047} \\
    & Recall & \textbf{0.6129} & \textbf{0.7646} & \colorcell{0.5915}{0.6129} & \colorcell{0.9563}{0.6129} & 0.0834 & 0.0740 & \colorcell{0.1098}{0.0834} & \textbf{\colorcell{0.8657}{0.0834}} \\
    & F1 score & \textbf{0.6458} & \textbf{0.5566} & \colorcell{0.6186}{0.6458} & \colorcell{0.9540}{0.6458} & 0.1309 & 0.0351 & \colorcell{0.1864}{0.1309} & \textbf{\colorcell{0.9132}{0.1309}} \\
\midrule
\multirow{3}{*}{\parbox{2cm}{RoBERTa}}  
    & Precision & 0.4359 & 0.0714 & \colorcell{0.7579}{0.4359} & \colorcell{0.9509}{0.4359} & 0 & 0 & \colorcell{0.5785}{0} & \textbf{\colorcell{0.9695}{0}} \\
    & Recall & 0.5 & 0.5 & \colorcell{0.6185}{0.5} & \textbf{\colorcell{0.9610}{0.5}} & 0 & 0 & \colorcell{0.1436}{0} & \colorcell{0.8414}{0} \\
    & F1 score & 0.4658 & 0.125 & \colorcell{0.6513}{0.4658} & \textbf{\colorcell{0.9549}{0.4658}} & 0 & 0 & \colorcell{0.2140}{0} & \colorcell{0.8988}{0} \\
\midrule
\multirow{3}{*}{\parbox{1.6cm}{LLaMa 3.1-8B-Instruct}}  
    & Precision & 0.8318 & 0.5754 & \colorcell{0.5701}{0.8318} & \colorcell{0.9260}{0.8318} & 0.4760 & \textbf{0.2369} & \colorcell{0.4798}{0.4760} & \colorcell{0.9639}{0.4760} \\
    & Recall & 0.5956 & 0.6286 & 0.5903 & \colorcell{0.9441}{0.5956} & \textbf{0.1933} & \textbf{0.2035} & \colorcell{0.2573}{0.1933} & \colorcell{0.8146}{0.1933} \\
    & F1 score & 0.6299 & 0.4002& \colorcell{0.5758}{0.6299} & \colorcell{0.9312}{0.6299} & \textbf{0.2530} & \textbf{0.1590} & \textbf{\colorcell{0.3252}{0.2530}} & \colorcell{0.8798}{0.2530} \\
\midrule
\multirow{3}{*}{\parbox{2cm}{Qwen 3-8B}}  
    & Precision & \textbf{0.8804} & 0.5822 & \textbf{\colorcell{0.8110}{0.8804}} & \colorcell{0.8726}{0.8804} & \textbf{0.5005} & 0.2222 & \colorcell{0.4460}{0.5005} & \colorcell{0.9034}{0.5005} \\
    & Recall & 0.5499 & 0.6199 & \textbf{\colorcell{0.6653}{0.5499}} & \colorcell{0.7921}{0.5499} & 0.0952 & 0.1878 & \textbf{\colorcell{0.2853}{0.0952}} & \colorcell{0.8504}{0.0952} \\
    & F1 score & 0.5606 & 0.3678 & \textbf{\colorcell{0.7073}{0.5606}} & \colorcell{0.8068}{0.5606} & 0.1539 & 0.1564 & \colorcell{0.3221}{0.1539} & \colorcell{0.8738}{0.1539} \\
\bottomrule
\end{tabular}
\begin{minipage}{0.95\linewidth}
\vspace{1ex}
\small\textit{Note:} All metrics are macro average values. Binary-SR/DR: binary classification (sufficient vs. deficient). Multi-DR: multi-label classification for deficient aspects. Column headers indicate train-test setups (e.g., R, R means Train: Real data, Test: Real data, R: Real data; S: Synthetic data; R+S: Real and Synthetic data). Cells are colored relative to the R,R baseline: shades of red indicate values larger than the baseline (deeper shades for greater ratios, e.g., >1.5x \textcolor{darkred}{dark red}); shades of blue indicate values smaller than the baseline (deeper shades for smaller ratios, e.g., <0.5x \textcolor{darkblue}{dark blue}). Bold values indicate the maximum in each respective column.
\end{minipage}
\end{table*}

Our findings reveal that SR maintain superior neutrality and professionalism compared to DR, which exhibit higher negativity and emotional bias. The release of ChatGPT has fundamentally transformed academic peer review, rapidly increasing AI-generated content across both reviewer categories. This ubiquitous adoption of AI tools necessitates urgent policy interventions to establish clear guidelines and disclosure norms for AI use in peer review, ensuring both efficiency and scholarly integrity are preserved throughout the review process.

\subsection{Evaluation of Deficient Review Detection Performance}

In this section, we evaluate the performance of various models on deficient review detection and critically assess the utility of synthetic data augmentation as a viable strategy for improving classification accuracy. We conducted two sets of experiments: a binary classification task to distinguish SR from DR, and a more granular multi-label task to identify specific categories of deficient reviews. Our evaluation leverages encoder models (BERT, SciBERT, RoBERTa) and open LLMs (LLaMa 3.1-8B-Instruct, Qwen 3-8B). The performance results are presented in Table~\ref{tab:model_performance}.

\subsubsection{Distributional Analysis of Synthetic vs. Real Data}

\begin{figure}[htbp]
\centering
\includegraphics[width=\linewidth]{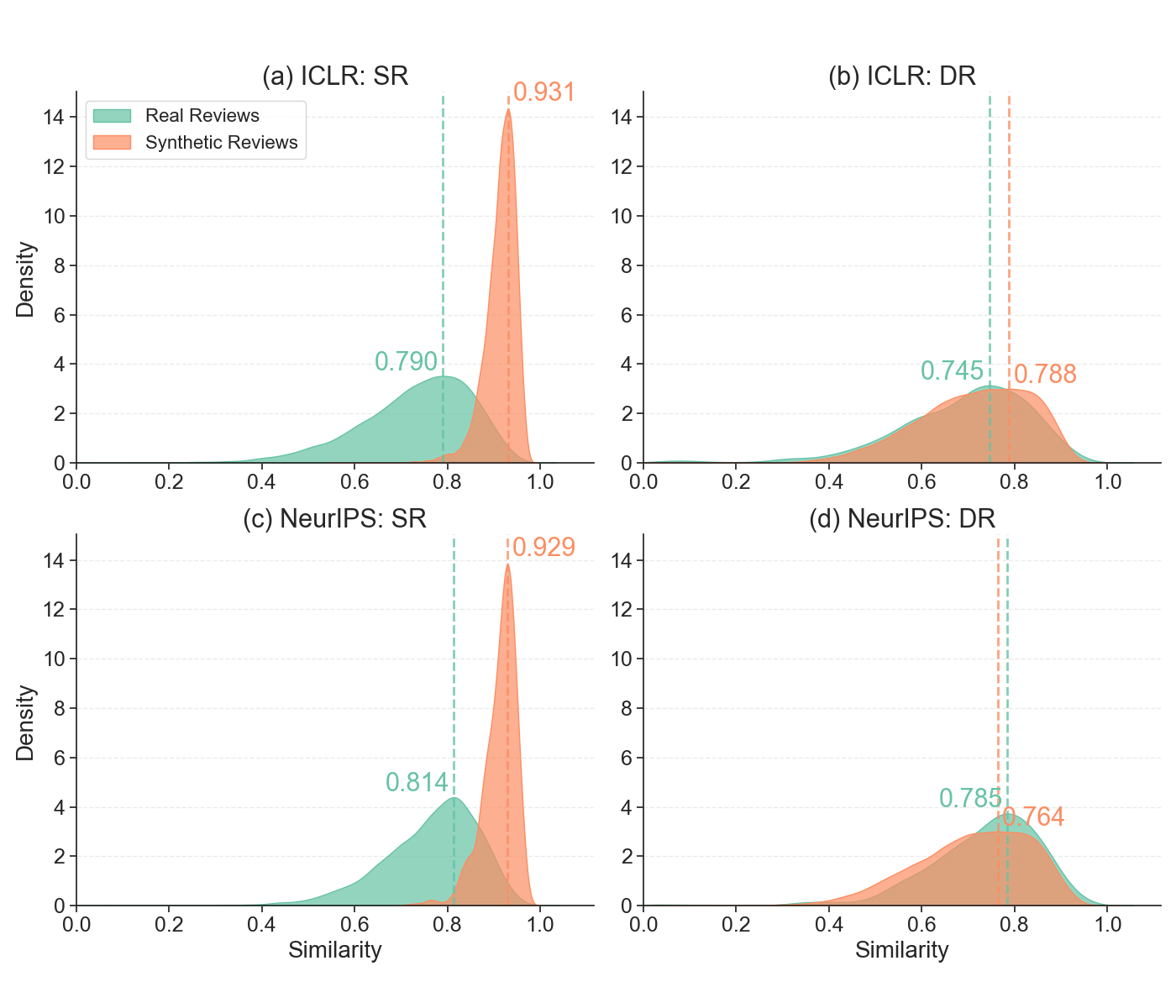}
\caption{Distribution of semantic similarity scores between paper abstracts and reviews, for both real and synthetic reviews from the ICLR and NeurIPS datasets. Dashed lines indicate peak similarity for each group.}
\label{review-compare}
\end{figure}

To establish a robust foundation for interpreting model performance, we first examine the semantic alignment between synthetic and real reviews. Understanding the distributional characteristics of the training data is essential for accurate model evaluation and generalization assessment. We employ Sentence-BERT (all-mpnet-base-v2) to measure semantic similarity between review content and the corresponding paper abstracts, uncovering substantial distributional differences between the real and synthetic datasets, as shown in Figure~\ref{review-compare}.

Synthetic SR exhibit significantly higher and more concentrated semantic similarity scores than their real counterparts. This distributional shift reflects our generation methodology, whereby GPT-4.1 produces reviews conditioned solely on abstract content, leading to a more homogeneous focus on paper summaries relative to human reviewers, who incorporate broader contextual knowledge and diverse rhetorical strategies. In contrast, DR show more comparable distributions between synthetic and real instances, indicating that both human- and model-generated deficient reviews display similar patterns of semantic divergence from core paper contributions. Compared with SR, the consistently lower similarity scores for DR across both datasets support defining deficient reviews as inadequate engagement with substantive scientific content.

\subsubsection{Baseline Performance and Generalization Challenges}

When models are trained and tested exclusively on real data (R, R), performance varies. For the binary SR/DR task, SciBERT, a model pre-trained on scientific text, achieves the highest baseline F1 score (0.6458). For the more challenging Multi-DR task, which involves identifying specific types of deficient, performance is substantially lower across all models, with LLaMa 3.1-8B-Instruct showing the best, albeit modest, F1 score (0.2530). This highlights the inherent difficulty of nuanced, multi-label deficient review detection with limited real-world data.

The performance drop in the (R, S) setting (training on real, testing on synthetic) confirms the impact of the distributional shift. For instance, SciBERT's F1 score on the binary task falls from 0.6458 to 0.5566, and LLaMa's drops from 0.6299 to 0.4002. This indicates that models trained on the patterns of real reviews struggle to generalize to the different, more ``idealized'' characteristics of the synthetic data, particularly the high-similarity sufficient reviews.

\subsubsection{The Efficacy of Synthetic Data Augmentation}

By training models on a composite dataset of real and synthetic examples and testing on a held-out set of real data (real + synthetic training, real testing; R+S, R), we uncover a nuanced relationship between data augmentation, task complexity, and model architecture.

\begin{itemize}
\item On the binary SR/DR task, the benefits of augmentation are conditional: mixed training with synthetic and real data yields substantial improvements in multi-label detection of deficient review types and enhances recall and F1 scores on the binary task across most models. For example, the strategy significantly boosts performance for models like Qwen 3-8B (F1 score improves from 0.5606 to 0.7073), yet it marginally hinders those already well-attuned to the domain, such as SciBERT and LLaMa 3.1-8B-Instruct. This finding serves as a crucial cautionary note: for simpler tasks and specialized models, augmenting with statistically divergent synthetic data can introduce counterproductive noise.

\item For the more complex Multi-DR task, synthetic data augmentation proves clearly advantageous. In this data-scarce scenario, augmentation delivers a marked performance boost across all models. For instance, LLaMa 3.1-8B-Instruct's F1 score improves from 0.2530 to 0.3252, while Qwen 3-8B's F1 score nearly doubles from 0.1539 to 0.3221. This result strongly indicates that synthetic data is a highly effective tool for addressing data scarcity in fine-grained classification problems, enabling models to learn more robust and generalizable representations of rare categories.
\end{itemize}

In summary, while synthetic data are not a perfect replica of real-world data, its strategic use as an augmentation tool holds promise, with effectiveness depending on task complexity and model architecture and proving most valuable for challenging, low-resource classification tasks.

\section{Conclusion}

We introduced ReviewGuard, an LLM-powered framework for detecting deficient peer reviews in machine learning conferences, revealing critical patterns that distinguish deficient from sufficient reviews: reduced structural complexity, lower ratings, heightened negativity, and a notable surge in AI-generated content following ChatGPT's release across all reviewer categories. We enhance model performance through fine-tuning detection models including encoder-based models and open-source LLMs with LLM-driven synthetic data augmentation, demonstrating significant efficacy improvements particularly in fine-grained multi-label deficient review detection. However, there are limitations of this study: first, the data augmentation is based on abstracts rather than full-text content, which may have bias in terms of knowledge representation. Second, data augmentation prompts also need to be improved the synthetic review's quality; for example, advanced algorithms like reinforcement learning could be implemented. Last but not the least, the current scope is limited to top ML/AI conferences, cross-validation in other domains needs to be conducted to further evaluate the effectiveness and generalizability of the proposed framework. In the future, we will extend ReviewGuard to large-scale, cross-disciplinary applications, incorporate multimodal data, and develop real-time deployment capabilities to establish robust AI-enhanced peer review systems to preserve academic integrity.

\section{Acknowledgments}
This research is partially supported by NSF grant \#2225229. The authors acknowledge the use of LLMs in this work. Specifically, LLMs were employed for data augmentation to expand the training dataset, and for editing and grammar enhancement during the manuscript preparation process. The core research contributions, methodology, analysis, and conclusions remain the original work of the authors.

\bibliographystyle{ieeetr}
\bibliography{references}

\end{document}